# Optimizing token usage on Large Language Model conversations using the Design Structure Matrix

Ramón María García Alarcia[1], Alessandro Golkar[1]

[1]Technical University of Munich

**Abstract:** As Large Language Models become ubiquitous in many sectors and tasks, there is a need to reduce token usage, overcoming challenges such as short context windows, limited output sizes, and costs associated with token intake and generation, especially in API-served LLMs. This work brings the Design Structure Matrix from the engineering design discipline into LLM conversation optimization. Applied to a use case in which the LLM conversation is about the design of a spacecraft and its subsystems, the DSM, with its analysis tools such as clustering and sequencing, demonstrates being an effective tool to organize the conversation, minimizing the number of tokens sent to or retrieved from the LLM at once, as well as grouping chunks that can be allocated to different context windows. Hence, this work broadens the current set of methodologies for token usage optimization and opens new avenues for the integration of engineering design practices into LLMs.

## 1 Introduction

The recent, rapid development and popularization of Large Language Models (LLM) have transformed the panorama of Natural Language Processing (NLP) and, more generally, of Artificial Intelligence (AI), permeating into society and transforming the way many tasks are performed, being now either supported or automated with the help of LLM-based tools.

Along with the challenges of hallucinations, lack of reasoning capabilities, inability to perform numerical calculations, natural aging of the training data, and improper traceability and citation of information sources, another intrinsic challenge of LLMs, tightly related to their architecture and training, concerns their limited context window and maximum token output (Kaddour et al., 2023).

Indeed, the context window is the cornerstone for LLM-based applications which require the previous interactions in the conversation to be preserved and considered by the LLM. This, being true for long conversations, is of particular importance in the engineering design field when an LLM is used to support engineers in the design of a system, going from high-level concept generation to lower-level system requirements or technical specifications generation. This application requires previous decisions as well as the decision-making process to be considered in later stages. Additionally, many applications require that large chunks of text be entered by the users or returned to them in single steps. In these cases, the token output limit can harm the performance of the system.

Although new LLMs with longer context windows and larger token output limits are slowly being released, they are still not a feasible solution for many individuals and organizations that need to adapt the LLMs to increase the performance in their target applications by means of additional training. The training and fine-tuning of long-context-window LLMs is significantly more computationally expensive.

And most importantly, many of the LLM-based applications rely on API calls to providers of LLMs, being billed not only taking into account the number of API calls, but also the number of tokens that are sent into the model and the number of tokens that are generated (and served to the user) by the model. Even in cases in which the LLM is run locally by the user, every token counts when considering power consumption and application speed. Thus, optimizing token usage remains a relevant point to be taken care of.

In this sense, the Design Structure Matrix (DSM) (Eppinger and Browning, 2012) is a promising tool for such cases of restricted token availability or required token usage optimization. Considering that many conversations can be understood as a process with a series of activities building one after the other, the DSM helps analyze the dependencies between the elements of the conversation and optimize it.

In this paper, we study the application of the DSM for the optimization of token usage in a case in which an LLM is used to support the design of a spacecraft, going from a high-level mission statement formulated in text by a customer to the generation of system requirements for the different spacecraft systems. We formulate a generic framework on how to apply the DSM, we apply it for our use case, and discuss the results. Results show a promising opportunity and a way forward for using the DSM as an additional tool for LLM token usage optimization.

## 2 State of the art

### 2.1 Large Language Models

Large Language Models (LLMs) have been around half a decade already, but it was not until the end of 2022 that gained widespread public interest when OpenAI released the ChatGPT interface, making their GPT-3 LLM (Brown et al., 2020) easily accessible to the masses. The technology behind LLMs, the transformer architecture, was introduced by Google in 2017, bringing together tokenization, word embeddings, and the novel attention mechanism (Vaswani et al., 2017) assigning different importance to different chunks of the processed text.

The results of the novel transformer architecture, with initial models based on it such as BERT (Devlin et al., 2019), quickly showed a significant performance increase with regards to the previous models used in Natural Language Processing (NLP) tasks and started a race to train ever-bigger models of billions of parameters, taking huge general corpora of data. GPT-4 from OpenAI (OpenAI et al., 2024), Gemini 1.5 Pro from Alphabet (Gemini Team et al., 2024), Llama 3 from Meta, Claude 3 Opus from Anthropic, or Mistral 8x7B by Mistral AI (Jiang et al., 2024), are some of the most prominent and advanced LLMs for the time being. As models become better at a variety of benchmarks, the focus is turning to developing their applications across industries, with fine-tuning, Retrieval Augmented Generation (RAG) (Lewis et al., 2021), and integration into broader tools being explored.

### 2.2 LLM conversations, tokens, context, and limitations

In essence, an LLM is a generative AI model that excels in text generation tasks by processing natural language and predicting the next word in a sequence with the highest probability. Thus, the natural way LLMs can be used is in conversations with a human user. LLMs work with tokens. Tokens are chunks of text that the model can process or generate. In the process of tokenization, natural language words are reduced to their constituent parts, which bring meaning, and these are transformed into a number or ID (i.e., "the token"). It is relevant to mention that there are many different tokenizers and related techniques. We use the state-of-the-art GPT 3.5/GPT 4 tokenizer from OpenAI in this work. As an approximation, one token roughly corresponds to one natural language word, but this is not an exact correspondence.

In the frame of a conversation, it is necessary that previous interactions between the language model and the user are remembered and then processed each time new responses are generated by the model. These previous interactions are stored and become a part of what is called the context window. However, the context window of an LLM and also its responses are limited by a maximum number of tokens, with context windows normally having a bigger token limit than LLM outputs. These token limits come from the architecture of the LLM themselves (i.e., the size and number of the model layers) and are hard to overcome as LLMs with large context windows are also computationally more expensive to train and require corpora of data with longer text sequences, which are harder to find.

Limitations in number of tokens have fueled research on context window extension methods (Peng et al., 2023) (Zhu et al., 2024), as well as methods that are rather directed toward the minimization of token usage (Li et al., 2023). To the best of our knowledge, this is the first academic work that proposes the usage of the Design Structure Matrix as a tool to optimize conversations and reduce the number of tokens that are used.

Table 1: Most prominent LLM models, accessibility, and context window and output sizes in tokens

| Model | Provider | Accessibility | Context window | Max. output tokens |
|---|---|---|---|---|
| gpt-4-turbo | OpenAI | Proprietary | 128 000 | 4 096 |
| gpt-4-32k | OpenAI | Proprietary | 32 768 | 4 096 |
| gpt-4 | OpenAI | Proprietary | 8192 | 4 096 |
| gpt-3.5-turbo | OpenAI | Proprietary | 16 385 | 4 096 |
| Gemini 1.5 Pro | Google | Proprietary | 128 000 | 8 192 |
| Gemini 1.0 | Google | Proprietary | 32 000 | 2 048 |
| Claude 3 family | Anthropic | Proprietary | 200 000 | 4 096 |
| Claude 2 | Anthropic | Proprietary | 100 000 | 4 096 |

Ramón María García Alarcia, Alessandro Golkar

| Llama 3 family | Meta | Open source | 8 192 | N/A |
|---|---|---|---|---|
| Llama 2 family | Meta | Open source | 4 096 | 4 096 |
| Mistral 8x22B | Mistral AI | Open source | 64 000 | N/A |
| Mistral 8x7B | Mistral AI | Open source | 32 000 | N/A |
| Mistral 7B | Mistral AI | Open source | 32 000 | 8 192 |
| Falcon family | Technology Innovation Institute | Open source | 2 048 | 2 048 |

### 2.3 The Design Structure Matrix

The Design Structure Matrix (DSM) is a tool coming from the engineering design discipline that allows one to visualize the elements that make up a system or a process, their interrelations or interactions, and, more generally, the global architecture of the system or process. A DSM takes the form of an N x N matrix where the N elements of the system or process are represented together with the connections between them. The matrix can be binary, if only the existence or absence of connection is represented, or numerical, whenever the strength of such connection is numerically depicted. (Eppinger and Browning, 2012)

A DSM can be built after analyzing and decomposing a system or a process into its fundamental components and then documenting (and maybe quantifying the intensity of) the connections between them, to afterward apply algorithms allowing the detection of patterns and rearranging components for a system or process optimization. DSMs can be used to model both static architectures (e.g., a product or an organization) and temporal flows (e.g., a process). In a process architecture DSM, elements are activities that have interfaces between them. Design can be thought as an iterative process (Kline, 1985), in which iterations are an important part of it. Iterations can be depicted as bidirectional connections between activities of the design process.

In this sense, clustering and sequencing algorithms applied to a design process DSM can help optimize the design process, and in the case in which this design process is happening in the frame of a conversation with an LLM, it can help optimize the conversation itself, and reduce the usage of tokens. Clustering allows us to find groups of activities that are particularly intertwined and group them together so that they can be sent to an LLM (or received back from the LLM) in a single chunk of input/output or even in a single conversation. Sequencing is a simple reordering technique of the DSM that aims to minimize interactions by moving the interaction marks closer to the diagonal of the matrix. By rearranging a conversation in such a way, the chances that an overflow of a context window eliminates information that is needed in the current step are reduced, since that information will be closer to the current point of the conversation.

## 3 Methodology

### 3.1 Use case and utilization flow

The use case for this paper is the design of spacecraft. A design assistant is built, which, based on an LLM and with the support of an Object-Process Methodology (OPM) (Dori, 2002) model of a generic spacecraft, iteratively selects parameters for different parameters of the subsystems, starting from a high-level mission statement written by the user in natural language.

In this use case, the design of a spacecraft is taken as an iterative process that can be modeled by a DSM, where the design activities correspond to the design of the different subsystems that a spacecraft can contain. The design process is realized in the frame of a conversation with an LLM. The conversation starts with the user specifying the mission statement for the spacecraft to be designed. Then, in this conversation, pieces of the OPM model of the spacecraft are exchanged between the user and the LLM in both their generic version (provided by the user) and the specific version containing a detailed design (returned by the LLM).

Without a DSM and its analysis methods of clustering and sequencing, the flow of the spacecraft design is as follows:

1. The OPM model is translated into text
2. A certain mission statement is entered by the user into the LLM
3. The full generic OPM model of a spacecraft is sent to the LLM

4. The LLM is asked to perform the appropriate design decisions and return an information-filled OPM
5. The LLM retrieves a full information-filled OPM model of the spacecraft, specifying a particular design

In this case, however, the following drawbacks might occur:

- The combination of the initially given mission statement and the full OPM model in its text version (including the generic version sent to the LLM and the information-filled version generated by it) might not fit into the context window size. As a result, the initial portions of the conversation will be progressively lost, and that information will not be considered when making new design choices.
- The information-filled OPM model might not completely fit inside the output size of the LLM, and as a result, the retrieved design will be incomplete and will not cover all the requested spacecraft subsystems or aspects. This is also a case in which tokens are lost, since they were provided by the user to the LLM, but the corresponding counterpart was not returned by the LLM.

The flow, when incorporating the DSM, can be depicted in the following manner:

1. The OPM model is translated into text
2. A DSM is created from the OPM model
3. Clustering is performed on the DSM
4. Sequencing is performed on the clustered DSM
5. The clusters are identified, and the corresponding chunks of text are isolated
6. Each chunk of text resulting from the clustering is sent to the LLM
7. The LLM returns the same chunk of text but filled with information coming from its design choices

In this way, thanks to the clustering, smaller pieces of conversation are sent and retrieved from the LLM at each step, making it easier for the output to be complete and cover the requested information. A token budget can be used to quantify the improvement of using the DSM to optimize the conversation flow. It is a simple but powerful tool that gives a clear insight before even deploying the conversation into an LLM and measuring the results and costs.

### 3.2 From an OPM space mission model to LLM tokens and to a DSM representation

An Object-Process Methodology model can be easily expressed in natural language thanks to the native bimodal representation of a visual Object-Process Diagram (OPD) with its text equivalent in Object-Process Language (OPL). Figure 1 depicts the OPD visualization of the highest level of the space mission OPM model as well as a lower-level subsystem of the space mission's spacecraft. Figure 2 depicts a portion of the OPL representation of the OPM model.

Additionally, an Object-Process Methodology model can also be expressed in the form of a Design Structure Matrix. Existent algorithms already perform this operation and are embodied in OPM drafting tools such as OPcloud and OPCAT (Dori et al., 2010). Despite this, for this work, a manual version of the DSM at a higher level is created, containing only the following general parameters and subsystems that are available in the OPM model and that, in the frame of the conversation, are considered activities of the design process:

- Mission Statement
- Payload
- Orbit
- Telemetry and Telecommand
- Ground Station
- Attitude and Orbit Control System
- Propulsion
- Electrical and Power System
- Thermal Control System
- Spacecraft's general parameters
- Structure
- Launcher

The interactions between systems are derived from the experience and references in spacecraft design, such as (Larson and Wertz, 1992). The strength of the interactions corresponds to the number of tokens that are generated when tokenizing the OPL text corresponding to each of the OPM model portions.

Ramón María García Alarcia, Alessandro Golkar

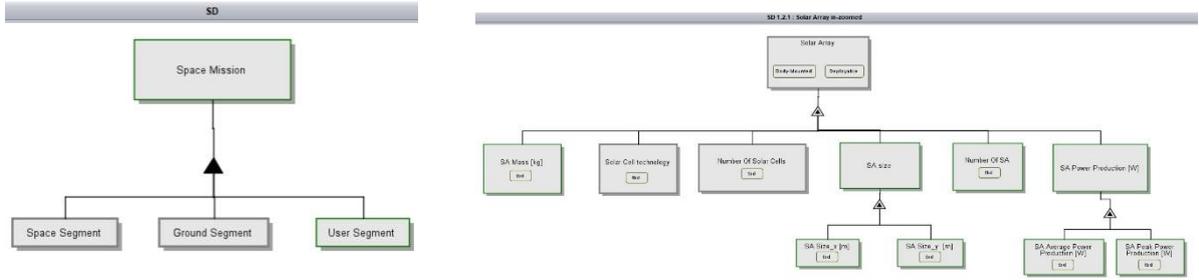

Figure 1: OPD visualization of the highest level of the space mission's OPM model (left). OPD visualization of the Solar Array subsystem present in a spacecraft. *Tbd*'s mark pieces of information to be completed by the LLM taking design choices (right).

Figure 2: Portion of the OPL text representation of the OPM space mission model.

### 3.3 Clustering and sequencing algorithms with the DSM

Clustering can be described mathematically as an optimization problem. In it, activities are grouped together into clusters. The goal is to find the grouping that minimizes a cost function J. The cost function considers both the summation of the sizes of clusters $C$ and the number of interactions between clusters $I\_0$, as defined per Equation 1 below (Eppinger and Browning, 2012).

Equation 1: Clustering minimization objective function (Eppinger and Browning, 2012).

$$\min J = \min \alpha \sum_{i=1}^{M} C_i^2 + \beta I_0$$

$\alpha$ and $\beta$ are the weighting parameters for the summation of the size of the clusters and the number of interactions between clusters, respectively. For our application, we desire $\alpha > \beta$ as the main purpose is to reduce the size of the chunks that are sent to or retrieved from the LLM. Clustering is implemented on Julia using the *DesignStructureMatrix.jl* package implementing the algorithm referenced in (Damasio et al., 2017). Sequencing is a reordering operation performed on a DSM to avoid feedback marks. There are different methods to do so, and in this work, a method based on a *reachability matrix* is used. The algorithm first computes the so-called reachability matrix, which represents the intersection between a matrix with the input sets and the output sets for the elements of the matrix. Sequencing is implemented on Julia using the *DesignStructureMatrix.jl* package implementing the algorithm referenced in (Warfield, 1973).

### 3.4 Token budget

The token budget is a tool that allows us to assess whether a conversation is feasible, taking into account the size of the context window and the maximum output from an LLM. When the budget is negative, it is either because more tokens are required for the full conversation than the length of the LLM context window, or because at a given step, the requested output tokens exceed the output capability of the LLM. Additionally, the token budget can help calculate the costs of the conversation, especially when interfacing with external, API-based LLM services.

Table 2 shows the elements that need to be considered and tokenized to compute the Window Budget $WB$, and if they are an input to the LLM or an output from it. In order to compute the Window Budget, both input and outputs are considered, while to compute the Output Budget, only conversation outputs are considered:

Table 2: Elements of the conversation, codification, and input/output.

| Element | Code | Input | Output |
|---|---|---|---|
| User statement | $USt$ | X | |
| Instructions to the LLM | $IntLLM$ | X | |
| Generic space mission model | $GM$ | X | |
| Information-filled space mission model | $FM$ | | X |

Additionally, a margin $MG$ is considered in the computation of the budgets to account for the uncertainty of both the user input and the results of the LLM inference. The Window Budget, for our conversation, is then computed as: $WB = (USt + IntLLM + GM + FM) \cdot MG$

The $WB$ is considered positive whenever it is bigger than the LLM's context window $CW$ size ($WB < CW$). The Output Budget is as follows for the conversation we set up: $OB = FM \cdot MG$

The $OB$ is positive whenever it is below the maximum number of output tokens $OL$ of the model ($OB < OL$).

## 4 Results

In order to understand the benefits of the methodology, we apply it following the sequential steps outlined. To begin with, we calculate an initial token budget. For this particular case, we use the *Mistral 7B* LLM from Mistral AI, being publicly available for download and running locally. The initial token budget is available in Table 3. As seen, in this case, the results are mixed: while the design conversation fits into the generous context window, the conversation needs to be optimized and split so that around 1400 tokens are not lost in outputs.

Table 3: Initial token budget before the DSM-driven conversation optimization.

| Item | Number of tokens |
|---|---|
| *SINGLE CONVERSATION PIECE* | |
| User statement - $USt$ | 200 |
| Instructions to the LLM - $IntLLM$ | 200 |
| Generic model - $GM$ | 9619 |
| Information-filled model - $FM$ | 9619 |
| Variability margin – $MG$ | 5% |
| **Window budget - $WB$** | **21 601** |
| *Context window Mistral 7B - CW* | *32 000* |
| **$CW - WB$** | **+ 10 399** |
| **Output budget - $OB$** | **9 619** |
| *Max. output Mistral 7B - OL* | *8 192* |
| **$OL - OB$** | **-1 427** |

Ramón María García Alarcia, Alessandro Golkar

In order to remediate this, we will use the DSM as explained in the methodology. First, as depicted in Table 4, a binary DSM is created from the space mission model in OPM, which depicts the interactions between the different systems and subsystems of the spacecraft, which are considered as activities of the design process, which happens in the design conversation. The DSM is read row-to-column. The design elements on the rows are influenced by the X-marked columns.

Table 4: Binary DSM corresponding to the OPM space mission model, created manually.

|  |  | A | B | C | D | E | F | G | H | I | J | K | L | M |
|---|---|---|---|---|---|---|---|---|---|---|---|---|---|---|
| **Mission S.** | A |  |  |  |  |  |  |  |  |  |  |  |  |  |
| **Payload** | B | X |  |  |  |  |  |  |  |  |  |  |  |  |
| **Orbit** | C | X |  |  |  |  |  |  |  |  |  |  |  |  |
| **AOCS** | D | X | X | X |  |  | X |  |  | X |  |  |  |  |
| **Propulsion** | E | X |  | X | X |  |  |  |  |  |  |  |  |  |
| **TT&C** | F |  | X | X |  |  |  | X | X |  |  |  |  |  |
| **GS** | G |  |  | X |  |  |  |  |  |  |  |  |  |  |
| **CDH** | H |  | X |  |  |  | X | X |  |  |  |  |  |  |
| **EPS** | I |  | X | X | X | X | X |  | X |  | X |  |  |  |
| **TCS** | J |  | X | X | X | X | X |  | X | X |  |  |  |  |
| **Structure** | K |  | X |  | X | X | X |  | X | X | X |  | X |  |
| **Spacecraft** | L | X | X |  | X | X | X |  | X | X | X |  |  |  |
| **Launcher** | M |  | X |  |  |  |  |  |  |  |  | X |  |  |

To perform proper clustering and sequencing operations, the DSM needs to be transformed from the binary form into a numerical one. Thus, each of the design elements, expressed in text in an OPL form, is tokenized, and the number of tokens is counted. As a first approximation, we assume the worst-case scenario in which all the tokens of a design element are influenced by the dependencies.

Table 5: Numerical DSM created when considering the tokens of the OPL portions.

|  |  | A | B | C | D | E | F | G | H | I | J | K | L | M |
|---|---|---|---|---|---|---|---|---|---|---|---|---|---|---|
| **Mission S.** | A |  |  |  |  |  |  |  |  |  |  |  |  |  |
| **Payload** | B | 190 |  |  |  |  |  |  |  |  |  |  |  |  |
| **Orbit** | C | 190 |  |  |  |  |  |  |  |  |  |  |  |  |
| **AOCS** | D | 190 | 272 | 3082 |  |  | 564 |  |  | 1416 |  |  |  |  |
| **Propulsion** | E | 190 |  | 3082 | 1082 |  |  |  |  |  |  |  |  |  |
| **TT&C** | F |  | 272 | 3082 |  |  |  | 497 | 1510 |  |  |  |  |  |
| **GS** | G |  |  | 3082 |  |  | 564 |  |  |  |  |  |  |  |
| **CDH** | H |  | 272 |  |  |  | 564 | 497 |  |  |  |  |  |  |
| **EPS** | I |  | 272 | 3082 | 1082 | 769 | 564 |  | 1510 |  | 60 |  |  |  |
| **TCS** | J |  | 272 | 3082 | 1082 | 769 | 564 |  | 1510 | 1416 |  |  |  |  |
| **Structure** | K |  | 272 |  | 1082 | 769 | 564 |  | 1510 | 1416 | 60 |  | 16 |  |
| **Spacecraft** | L | 190 | 272 |  | 1082 | 769 | 564 |  | 1510 | 1416 | 60 |  |  |  |
| **Launcher** | M |  | 272 |  |  |  |  |  |  |  |  | 239 |  |  |

The previous table is encoded into a matrix in Julia. The *DesignStructureMatrix.jl* package allows for clustering and sequencing of the matrix and then for simple visualization without the numerical weights. The results are depicted in Figure 3.

Optimizing token usage on Large Language Model conversations using the Design Structure Matrix

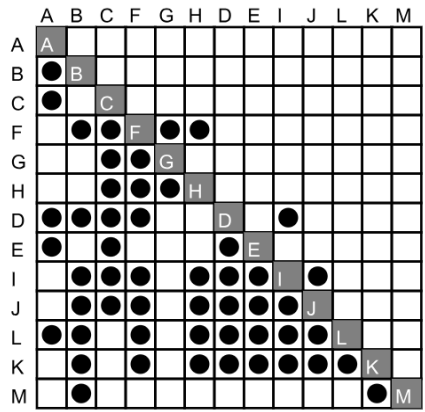

Figure 3: Results of the clustering and sequencing applied to the manually created DSM.

As depicted in Figure 3, the following sequence is obtained after applying sequencing:

1. Mission Statement (A)
2. Payload (B)
3. Orbit (C)
4. TT&C (F)
5. GS (G)
6. CDH (H)
7. AOCS (D)
8. PROP (E)
9. EPS (I)
10. TCS (J)
11. Spacecraft (L)
12. Structure (K)
13. Launcher (M)

Next, clusters are visually identified, as depicted in Figure 4, with red and blue squares.

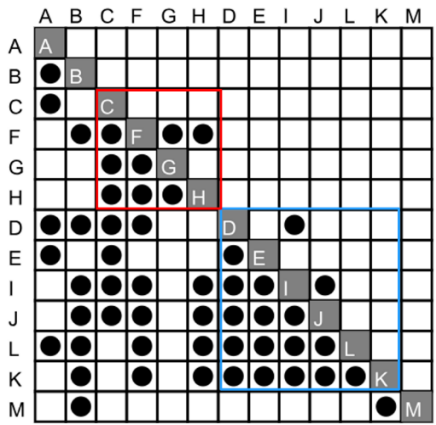

Figure 4: Visual identification of clusters in the DSM matrix.

From Figure 4, the following clusters are visually determined:

- *Cluster 1:* C, F, G, H - Orbit, TT&C, GS, CDH
- *Cluster 2:* D, E, I, J, L, K - AOCS, Propulsion, EPS, TCS, Structure

Taking into account the results of the sequencing and the identified clusters from the DSM analysis techniques, the proposed conversation when taking into account clustering and sequencing is the following:

- *1st conversation piece:* Mission Statement (A)+Payload (B) - 462 tokens
- *2nd conversation piece:* Orbit (C)+TT&C (F)+GS (G)+CDH (H) - 1438 tokens

Ramón María García Alarcia, Alessandro Golkar

- *3rd conversation piece:* AOCS (D)+PROP (E)+EPS (I)+TCS (J)+Spacecraft (L)+Structure (K) - 7703 tokens
- *4th (and last) conversation piece:* Launcher (M) - 16 tokens

When analyzing this new DSM-derived conversation sequence with a token budget, the following results are obtained:

Table 6: Final token budget after the usage of the DSM for conversation optimization.

| Item | Number of tokens |
|---|---:|
| *COMMON CONVERSATION PARAMETERS* | |
| Maximum output tokens - $OL$ | 8 192 |
| Context window - $CW$ | 32 000 |
| Instructions to the LLM - $IntLLM$ | 50 |
| Variability margin – $MG$ | 5% |
| *CONVERSATION PIECE 1* | |
| Mission statement - $USt$ | 190 |
| Generic model - $GM$ | 272 |
| Information-filled model - $FM$ | 272 |
| $OL - OB$ | **+7906** |
| *CONVERSATION PIECE 2* | |
| Generic model - $GM$ | 1438 |
| Information-filled model - $FM$ | 1438 |
| $OL - OB$ | **+6682** |
| *CONVERSATION PIECE 3* | |
| Generic model - $GM$ | 7703 |
| Information-filled model - $FM$ | 7703 |
| $OL - OB$ | **+103** |
| *CONVERSATION PIECE 4* | |
| Generic model - $GM$ | 16 |
| Information-filled model - $FM$ | 16 |
| $OL - OB$ | **+8175** |
| COMPLETE CONVERSATION (single context) | |
| $CW - WB$ | **+11 989** |

When compared to Table 5, an improvement in the budget is observed, directly speaking of the benefits of applying the DSM tool to analyze the conversation and optimize it by clustering it, sequencing it, and organizing it appropriately into different pieces. In cases of LLMs with less generous context windows, the conversation pieces can also be sent to separate conversation sessions with a lower loss of information -repeating, for instance, the indispensable information. In the cases in which only the output tokens are limiting, such as the one analyzed here, the methodology is still useful to avoid losses of tokens in the expected returns.

## 5 Conclusions

In this work, we presented a methodology for optimizing token usage in LLM conversations, using the Design Structure Matrix and its related sequencing and clustering techniques. This initial work, which, to the best of our knowledge, brings the DSM tool of the engineering design discipline to LLMs for the first time, shows preliminary results on a conversation session based on the design of a system (in this case, a spacecraft). The results show a promising path forward in incorporating DSM for LLM conversation analysis and optimization, albeit for now, use-case limited. Future work will center on the generalization of the methodology to a broader range of LLM conversations and improving certain aspects, such as removing the assumption of all tokens needing to be shared between the connected conversation elements, implementing clustering not requiring visual determination, or comparing different sequencing and clustering algorithms.

**Contact: Ramón María García Alarcia,** Technical University of Munich, Department of Aerospace and Geodesy, Lise-Meitner-Strasse 9, 85521, Ottobrunn, Germany, +49 89 289 55752, ramon.garcia-alarcia@tum.de, https://www.asg.ed.tum.de/en/sps/home/